\documentclass[11pt]{article}
\usepackage[letterpaper,margin=0.85in,top=0.8in,bottom=0.9in]{geometry}
\usepackage[T1]{fontenc}
\usepackage[utf8]{inputenc}
\usepackage{lmodern}
\usepackage{microtype}
\usepackage{amsmath,amssymb,mathtools}
\usepackage{graphicx}
\setkeys{Gin}{keepaspectratio}
\usepackage{xcolor}
\usepackage{booktabs}
\usepackage{array}
\usepackage{enumitem}
\usepackage{fancyhdr}
\usepackage{titlesec}
\usepackage[font=small,labelfont=bf]{caption}
\usepackage{float}
\usepackage{hyperref}
\usepackage{xurl}
\usepackage{fvextra}
\usepackage{placeins}
\usepackage{needspace}
\usepackage{etoolbox}

\hypersetup{
  colorlinks=true,
  linkcolor=black,
  citecolor=black,
  urlcolor=blue!45!black,
  pdfauthor={Christopher Da Silva},
  pdftitle={Hylos: Operability Contracts for Model-Native Spatial
Intelligence}
}
\setlist[itemize]{leftmargin=1.45em,itemsep=0.12em,topsep=0.22em}
\setlist[enumerate]{leftmargin=1.55em,itemsep=0.14em,topsep=0.22em}

\DefineVerbatimEnvironment{verbatim}{Verbatim}{breaklines=true,breakanywhere=true,fontsize=\scriptsize,frame=single,rulecolor=\color{black!18},framesep=2mm}
\DefineVerbatimEnvironment{Highlighting}{Verbatim}{breaklines=true,breakanywhere=true,fontsize=\scriptsize,frame=single,rulecolor=\color{black!18},framesep=2mm}

\newcommand{\tightlist}{\setlength{\itemsep}{0pt}\setlength{\parskip}{0pt}}

\titleformat{\section}{\Large\bfseries}{\thesection.}{0.6em}{}
\titleformat{\subsection}{\large\bfseries}{\thesubsection}{0.6em}{}
\titleformat{\subsubsection}{\normalsize\bfseries}{\thesubsubsection}{0.6em}{}
\titlespacing*{\section}{0pt}{1.15em plus 0.2em minus 0.1em}{0.42em}
\titlespacing*{\subsection}{0pt}{0.85em plus 0.2em minus 0.1em}{0.3em}
\titlespacing*{\subsubsection}{0pt}{0.65em plus 0.2em minus 0.1em}{0.25em}
\preto\section{\FloatBarrier\Needspace{6\baselineskip}}

\begin{document}

\begin{center}
{\LARGE \bfseries Hylos: Operability Contracts for Model-Native Spatial
Intelligence\par}
\vspace{0.55em}
{\large Christopher Da Silva\par}
\vspace{0.2em}
{\normalsize May 2026\par}
\end{center}
\vspace{0.3em}

\begin{abstract}
\noindent Foundation models can increasingly describe, reconstruct, and
generate 3D objects, assemblies, scenes, and environments, but visually
plausible spatial output is not yet operable 3D. A generated object,
product assembly, route, room, or environment becomes useful to an agent
only when the system can identify its entities, frames, surfaces,
constraints, provenance, operable handles, admissible actions, expected
effects, and validation failures. This paper argues that the missing
abstraction for spatial foundation models is an \emph{operability contract}:
a representation and runtime boundary that makes generated, imported, or
edited physical 3D inspectable, modifiable, validated,
provenance-backed, and transaction-safe.

We introduce Hylos, a systems architecture for contract-bounded spatial
intelligence. Hylos maintains scene-scale operability state over objects,
assemblies, assets, surface anchors, assertions, action candidates,
solver jobs, shared actuator invocations, spatial marks, work artifacts,
capability gaps, and effect diffs. Durable spatial changes are routed
through a \texttt{SpatialTransaction}: a commit boundary that resolves
references, checks admissibility, protects invariants, projects effects,
and returns commit, review, rollback, deferral, or capability-gap
outcomes.

The paper is framed as a systems/position preprint with a focused public
artifact study rather than a broad benchmark paper. The artifact study
examines causal repair: a visible misalignment appears on a dependent
component, while the supported repair lies upstream in the placement
structure that controls that component. The successful interaction traces
the symptom through scene dependencies, selects a supported upstream
interaction, and applies a validated change instead of directly editing
visible geometry.

We then generalize the architecture into a staged research program for
model-native spatial artifacts. Current systems select bounded graph
operations over explicit operability state; transitional systems wrap
meshes, splats, scans, and neural assets with recovered structure,
uncertainty, and provenance; future systems may co-generate geometry,
topology, constraints, handles, and audit hooks as a single operable
artifact. The central claim is that spatial AI should be evaluated not
only by visual quality, but by whether generated or edited 3D can become
a reliable substrate for CAD, robotics, simulation, inspection,
manufacturing, and interactive world authoring. The paper therefore frames
Hylos as a systems architecture and research program: transaction-safe
spatial state today, followed by guarded geometric/symbolic generation,
latent repair, parametric neural edit handles, and self-supervised
operability loops for model-native spatial artifacts.
\end{abstract}

\section{Introduction}\label{introduction}

Spatial AI is often framed as a generation problem: can a model produce
a convincing object, product assembly, mesh, room, video-consistent
world, or neural field? That framing is necessary but incomplete. Agents
do not merely look at space. They must inspect objects, reason over
parts and assemblies, route through environments, simulate consequences,
fabricate from geometry, validate changes, and recover when a proposed
change violates a physical or operational constraint.

This creates an \textbf{operability gap}. A generated spatial artifact
may look correct while still failing to answer basic operational
questions:

\begin{itemize}
\tightlist
\item
  Which entities, frames, surfaces, regions, and boundaries exist?
\item
  Which claims are measured, inferred, user-authored, or unsupported?
\item
  Which parts of the scene may legally change?
\item
  Which invariants must survive a mutation?
\item
  Which effects were expected, realized, unexpected, or unchecked?
\item
  Which views can be projected to rendering, CAD, simulation, robotics,
  or audit systems?
\end{itemize}

Without answers to these questions, a model-generated object or world
remains closer to a visual artifact than an agent-operable substrate. A
robot planner cannot safely route through a scene because an image
``looks clear.'' A CAD backend cannot certify a product because a mesh
``looks manufacturable.'' A human collaborator cannot trust an edit
unless the system can explain what changed, why it was allowed, and what
evidence supports the result.

Hylos addresses this gap by treating spatial state as a transaction-safe
operable substrate. The model is not treated as a raw geometric kernel.
It is treated as an interpreter over structured alternatives: candidate
physical interpretations, admissible operations, evidence-supported
values, unresolved assumptions, and reviewable goals. Geometry-changing
actions are accepted only when they pass through a runtime contract.

The central claim is:

\begin{verbatim}
Spatial foundation models become more reliable collaborators when generated, imported, or
  edited physical 3D is mediated by an operability contract: typed scene state, evidence,
  admissible actuators, protected invariants, effect diffs, and transaction-safe commit
  semantics.
\end{verbatim}

This paper is both an architecture proposal and a staged research
agenda. The near-term architecture is deliberately conservative: bounded
graph operations, validation, rollback, and evidence-grounded reasoning.
The long-term goal is more ambitious: model-native spatial artifacts
that co-generate geometry, topology, constraints, edit handles, and
provenance, then enter the same transaction runtime for validation and
projection.

\subsection{Contributions}\label{contributions}

This paper makes six contributions:

\begin{itemize}
\tightlist
\item
  It formulates \textbf{spatial operability contracts} as a missing
  abstraction between visual 3D generation, scene graphs, CAD kernels,
  robotics planners, simulation engines, and agentic model interfaces.
\item
  It introduces \texttt{SpatialTransaction} as a public architectural
  primitive for contract-bounded spatial mutation: type it, reference
  it, validate it, diff it, and commit it.
\item
  It describes a shared actuator model in which human gestures, model
  proposals, backend candidates, and imported evidence converge on the
  same scene-truth commit path.
\item
  It formalizes the runtime as a graph-state transition system with
  explicit validation checks, protected invariants, capability gaps,
  and effect-diff semantics.
\item
  It presents a causal repair artifact study showing that a model can
  route a visible symptom through dependency structure to a validated
  upstream interaction instead of applying a local visual edit.
\item
  It proposes a staged roadmap from current transaction-safe graph
  operations to structure-wrapped neural assets, hybrid co-generation,
  and future model-native spatial artifacts with token-level guardrails
  and parametric neural edit handles.
\end{itemize}

\subsection{Scope Of Claims}\label{scope-of-claims}

This paper does not claim that current foundation models can already
generate physically certified, model-native 3D worlds. It also does not
report a broad quantitative benchmark across many environments. Its
public empirical anchor is a causal repair family designed to test
whether a prototype agent follows dependency structure rather than
applying a local visual edit. The implemented prototype substrate is
broader than that single public anchor: it includes scene-scale
operability state, action search, review and deferral paths, effect
tracking, and internal fixtures covering additional mutation,
frame-transform, support-region, multi-region consequence, and
variant-generation scenarios.

The broader claim is architectural and scientific: visual plausibility
is insufficient as the target for spatial intelligence. A useful spatial
model must produce or interact with objects, assemblies, and
environments that are operable by other systems. The paper uses the
repair study as a focused public artifact, while the existing Hylos
substrate already exercises a broader family of authoring, mutation,
recovery, projection, and consequence-aware interaction paths that
motivate the evaluation program for model-native spatial artifacts.

\begin{figure}[tbp]
\centering
\includegraphics[width=0.92\linewidth]{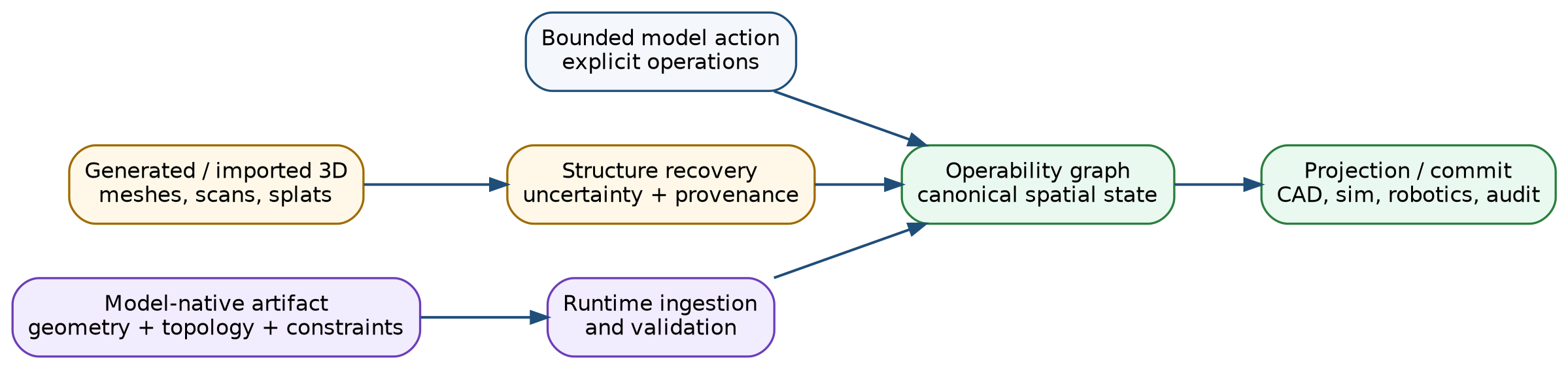}
\caption{Operability contract continuum. Current models select bounded
graph operations; transitional systems wrap neural assets with recovered
structure; future systems emit candidate model-native artifacts that
still pass through runtime ingestion and validation.}
\end{figure}

\section{The Operability Gap In Generated
3D}\label{the-operability-gap-in-generated-3d}

Generated 3D systems increasingly produce visually rich assets: meshes,
neural radiance fields, Gaussian splats, embodied environments, and
video-consistent worlds. These outputs expand the perceptual and
creative surface of spatial AI. However, a spatial artifact can be
visually impressive while remaining operationally inert.

An operable spatial artifact must expose more than shape. It must expose
the structure needed for agents and downstream systems to answer:

\begin{verbatim}
What exists?
What supports this claim?
What may change?
What must remain true?
What will this action affect?
What changed after execution?
What could not be checked?
\end{verbatim}

This matters because spatial interaction is causal and constructive. A
visible problem on object A may be caused by a relation to object B, a
frame on object C, or a constraint on assembly D. A creation request may
imply new surfaces, regions, attachments, openings, clearances, or
affordances that are not explicitly enumerated. Natural language often
collapses these distinctions:

\begin{verbatim}
"The receiving assembly looks laterally wrong relative to the body.
Fix the physical placement."
\end{verbatim}

For repair, the request identifies a symptom and a mission, but not
necessarily the correct interaction path. A robust spatial agent must
determine whether to move the visible dependent object, retarget a
relationship, adjust an attachment constraint, correct an upstream
frame, use an evidence-supported value, or defer because the available
evidence is insufficient.

For authoring, the analogous request may be:

\begin{verbatim}
"Create an accessible dispensing interface with a removable receiving tray."
\end{verbatim}

The agent must determine which topology, relationships, surfaces,
clearances, and attachment affordances are required, which construction
paths are permitted, and which realization constraints apply.

\subsection{Why Prompt Heuristics Are Not
Enough}\label{why-prompt-heuristics-are-not-enough}

It is tempting to solve such tasks with prompt rules such as ``center
the receiving tray'' or ``attach the component below the opening.''
These rules may pass a narrow fixture but do not scale. They also hide
spatial reasoning in prose, making it difficult to audit whether the
model made a supported inference or merely followed a task-specific
hint.

The boundary proposed here is different:

\begin{verbatim}
runtime: expose scene facts, evidence, legal targets, candidates, effects, and validators
model: propose and rank physical interpretations or construction goals
transaction kernel: accept, reject, review, or defer through explicit checks
\end{verbatim}

The model may infer. The runtime decides whether that inference is
admissible, grounded, and safe to commit.

\section{Related Work And
Positioning}\label{related-work-and-positioning}

This work sits at the intersection of semantic scene representation,
embodied-agent environments, tool-using language models, programmatic
geometry, and generated 3D objects, assemblies, and worlds. Its claim is
not that any one area is missing entirely. The claim is that agents need
an operability layer between semantic perception, visual generation, and
geometric execution.

\subsection{Semantic And Dynamic Scene
Graphs}\label{semantic-and-dynamic-scene-graphs}

3D scene graph work has shown that semantic spatial structure can unify
objects, rooms, cameras, and relationships in a 3D representation
{[}1{]}. Dynamic scene graph systems such as Kimera extend this idea
toward robot perception, mapping metric-semantic structure across time
and supporting planning over hierarchical spatial abstractions {[}2{]}.
These systems motivate the central representation choice here: spatial
reasoning should operate over explicit entities and relations, not only
over images, meshes, or text descriptions.

Hylos differs in emphasis. Existing scene graph work primarily asks how
to represent or reconstruct the world. This paper asks how an agent can
safely interact with such a representation: which variables are legal to
change, which values or constructions are supported, which relations
explain effects, and which checks must pass before a model's
interpretation becomes committed spatial state.

\subsection{Agentic Language Models And Tool
Use}\label{agentic-language-models-and-tool-use}

ReAct demonstrated that language models can interleave reasoning and
acting, using external observations to update plans {[}3{]}. Toolformer
showed that models can learn when and how to call tools {[}4{]}. In
robotics, SayCan grounds language-model proposals in feasible robot
affordances {[}5{]}, while Code as Policies uses language models to
generate executable policy code over robot APIs {[}6{]}.

These works establish the importance of tool-mediated action. Hylos
applies a related principle to semantic-spatial state: the model may
reason, but it must interact through typed scene contracts, legal
mutation or construction paths, value affordances, compiler validation,
and backend audit. The ``tool'' is not merely a callable API. It is an
operable spatial world model with enforceable invariants.

\subsection{Programmatic Geometry And Structured Shape
Generation}\label{programmatic-geometry-and-structured-shape-generation}

Programmatic 3D systems such as ShapeAssembly represent shape structure
as editable programs rather than as unstructured geometry {[}7{]}. This
supports a key premise of the present work: structured construction is
more controllable and inspectable than direct mesh generation. However,
programmatic shape generation by itself does not solve the agentic
interaction problem. An agent still needs to decide which operation is
intended, what supports it, which existing relation it affects, and
whether the resulting scene can be realized.

Hylos treats compiler-backed geometry as one execution substrate for
semantic decisions, not as the whole future of spatial intelligence. The
model is not the geometric kernel. It selects among legal
interpretations, values, and construction goals that are then lowered,
validated, projected, or rejected.

\subsection{Generated 3D Worlds And Spatial
Intelligence}\label{generated-3d-worlds-and-spatial-intelligence}

Large-scale embodied environments and asset corpora such as ProcTHOR
{[}8{]}, Objaverse {[}9{]}, and Holodeck {[}10{]} have expanded the
availability of interactive 3D worlds, objects, and generated
environments. Industry systems such as World Labs' Marble frame the
frontier as spatial intelligence: models that can perceive, generate,
reason about, and interact with 3D worlds {[}11{]}.

This paper is complementary to that direction. Generative world models
can produce rich spatial content. Operability contracts make such
content usable by agents. The contribution here is not another 3D
generator, but a runtime and artifact contract for creating, inspecting,
mutating, repairing, validating, and projecting structured spatial
state.

\subsection{Prototype Context}\label{prototype-context}

The experiments described here use Hylos, a prototype semantic-spatial
runtime and compiler for agentic interaction with structured 3D scene
state {[}12{]}. Hylos maintains semantic entities, derived causal views, relation/effect
claims, legal interaction paths, value affordances, compiler lowering
rules, backend realization, and audit outputs. This public draft
describes the prototype at the level needed to understand the claim and
evaluation, while omitting implementation-specific contracts and
internal mechanics.

\section{Hylos Runtime Contract}\label{hylos-runtime-contract}

Hylos treats spatial state as an operable graph plus a transaction
runtime. The graph records what the system believes about the scene. The
transaction runtime governs how that belief may change.

The core invariant is:

\begin{verbatim}
No spatial output becomes scene truth until Hylos can type it, reference it,
validate it, diff it, and commit it.
\end{verbatim}

This invariant applies equally to user gestures, model-authored
proposals, backend-generated candidates, imported evidence, generated
geometry, and future model-native spatial artifacts.

\subsection{Formal Operability State}\label{formal-operability-state}

Let the runtime state at step \(t\) be:

\[\mathcal{S}_t = \left(\mathcal{G}_t, \mathcal{E}_t, \mathcal{A}_t, \mathcal{C}_t, \mathcal{I}_t, \mathcal{R}_t, \mathcal{K}_t\right)\]

where:

\begin{itemize}
\tightlist
\item
  \(\mathcal{G}_t\) is the scene operability graph of entities, regions,
  boundaries, anchors, frames, assertions, and relations;
\item
  \(\mathcal{E}_t\) is the evidence set, including measurements, user
  declarations, sensor observations, model proposals, and imported
  references;
\item
  \(\mathcal{A}_t\) is the set of admissible actuator invocations
  available in the current state;
\item
  \(\mathcal{C}_t\) is the set of capability gaps and unresolved
  requirements;
\item
  \(\mathcal{I}_t\) is the set of protected invariants that must survive
  a transaction;
\item
  \(\mathcal{R}_t\) is the set of realization projections into display,
  CAD, simulation, robotics, or audit views;
\item
  \(\mathcal{K}_t\) is the runtime knowledge about validators, lowerers,
  candidate generators, and projection adapters.
\end{itemize}

A spatial change is not modeled as direct geometry mutation. It is
modeled as a guarded state transition:

\[\mathcal{T}: \mathcal{S}_t \rightarrow \mathcal{S}_{t+1} \cup \{\text{review}, \text{rollback}, \text{capability gap}\}\]

The transition may commit a new state, request human review, roll back
to the prior state, or record an explicit missing capability. This
framing treats failure as structured information rather than as an
incidental exception.

\subsection{Scene Operability Graph}\label{scene-operability-graph}

The scene operability graph is the canonical spatial state. It is not a
renderer object list and not a product-specific part tree. It records
evidence-backed spatial-physical structure:

\begin{itemize}
\tightlist
\item
  entities, surfaces, anchors, regions, boundaries, and coordinate
  frames;
\item
  assertions, constraints, affordances, and unresolved hypotheses;
\item
  operation candidates, validation results, and capability gaps;
\item
  provenance linking claims to user input, measurements, imported
  evidence, or model proposals;
\item
  realization artifacts such as meshes, CAD outputs, splats, or
  simulation projections.
\end{itemize}

Downstream rendering, CAD, simulation, robotics, and audit views are
projections of this state. They may realize or display valid subsets of
the graph, but they do not define scene truth.

\subsection{Derived Causal Views Over Operability State}\label{derived-causal-views-over-operability-state}

The scene operability graph records canonical spatial state. For
transaction-time reasoning, Hylos derives causal views over this state:
variables, claims, constraints, dependencies, unresolved assumptions,
provenance, and validation status. These causal views are projections
over operability state rather than separate persistent task graphs.

This distinction prevents the runtime from fragmenting into
product-specific structures such as route graphs, access graphs,
affordance graphs, repair graphs, or interaction graphs. A claim such as
support, obstruction, access, containment, alignment, or occlusion may
appear as a recognized normalized claim kind, but the contract does not
require a closed ontology of every possible spatial predicate.

Instead, the runtime asks generic dependency questions: which references
are involved, what effect is intended, which variables may change, which
constraints must be preserved, which dependencies are satisfied,
missing, violated, or uncertain, and what evidence supports each claim.

The same derived causal view can support repair, placement, route
planning, authoring, inspection, artifact ingestion, and model-generated
scene updates while preserving the main invariant: unsupported claims
remain unresolved knowledge, review items, or capability gaps rather than
becoming fake geometry or renderer-local state.

\subsection{Spatial Transactions}\label{spatial-transactions}

A \texttt{SpatialTransaction} is the durable commit boundary for a
spatial change. It wraps a proposed mutation in a verification envelope:

\begin{verbatim}
context certificate
-> typed operation or shared actuator invocation
-> precondition checks
-> protected invariants
-> derived graph mutation or realization patch
-> backend/audit result
-> effect diff
-> commit, review, rollback, or capability gap
\end{verbatim}

The transaction contract lets the system distinguish a healthy refusal
from an unhealthy workaround. A healthy refusal produces a typed missing
layer: missing measurement, missing legal target, missing value
acquisition, missing candidate, missing lowerer, missing backend
operation, or missing verification. An unhealthy workaround hides the
failure inside a prompt hint, regex branch, manual numeric nudge, or
renderer-only fallback.

Formally, a transaction can be represented as:

\[\mathcal{T} = \left(\mathcal{C}_{ctx}, \mathcal{O}, \mathcal{P}_{pre}, \mathcal{I}_{prot}, \mathcal{V}, \mathcal{M}, \mathcal{B}, \Delta\mathcal{E}\right)\]

where \(\mathcal{C}_{ctx}\) is the context certificate, \(\mathcal{O}\)
is the typed actuator invocation or graph operation,
\(\mathcal{P}_{pre}\) is the precondition set, \(\mathcal{I}_{prot}\) is
the protected invariant set, \(\mathcal{V}\) is the validator family,
\(\mathcal{M}\) is the derived graph mutation or compatibility patch,
\(\mathcal{B}\) is the backend or projection audit result, and
\(\Delta\mathcal{E}\) is the effect diff.

A transaction is admissible only if:

\[
\begin{aligned}
\operatorname{Admit}(\mathcal{T}, \mathcal{S}_t) ={}& \operatorname{Legal}(\mathcal{O}, \mathcal{A}_t) \land \operatorname{Grounded}(\mathcal{O}, \mathcal{E}_t) \\
& \land \operatorname{Pre}(\mathcal{P}_{pre}, \mathcal{S}_t) \land \operatorname{Preserve}(\mathcal{I}_{prot}, \mathcal{S}_t, \mathcal{M}) \\
& \land \operatorname{Realizable}(\mathcal{M}, \mathcal{K}_t).
\end{aligned}
\]

If any predicate is false, the runtime must not lower raw text, infer
hidden geometry, or mutate renderer-local state. It must produce a typed
review item, rollback, or capability gap.

\begin{figure}[tbp]
\centering
\includegraphics[width=0.78\linewidth]{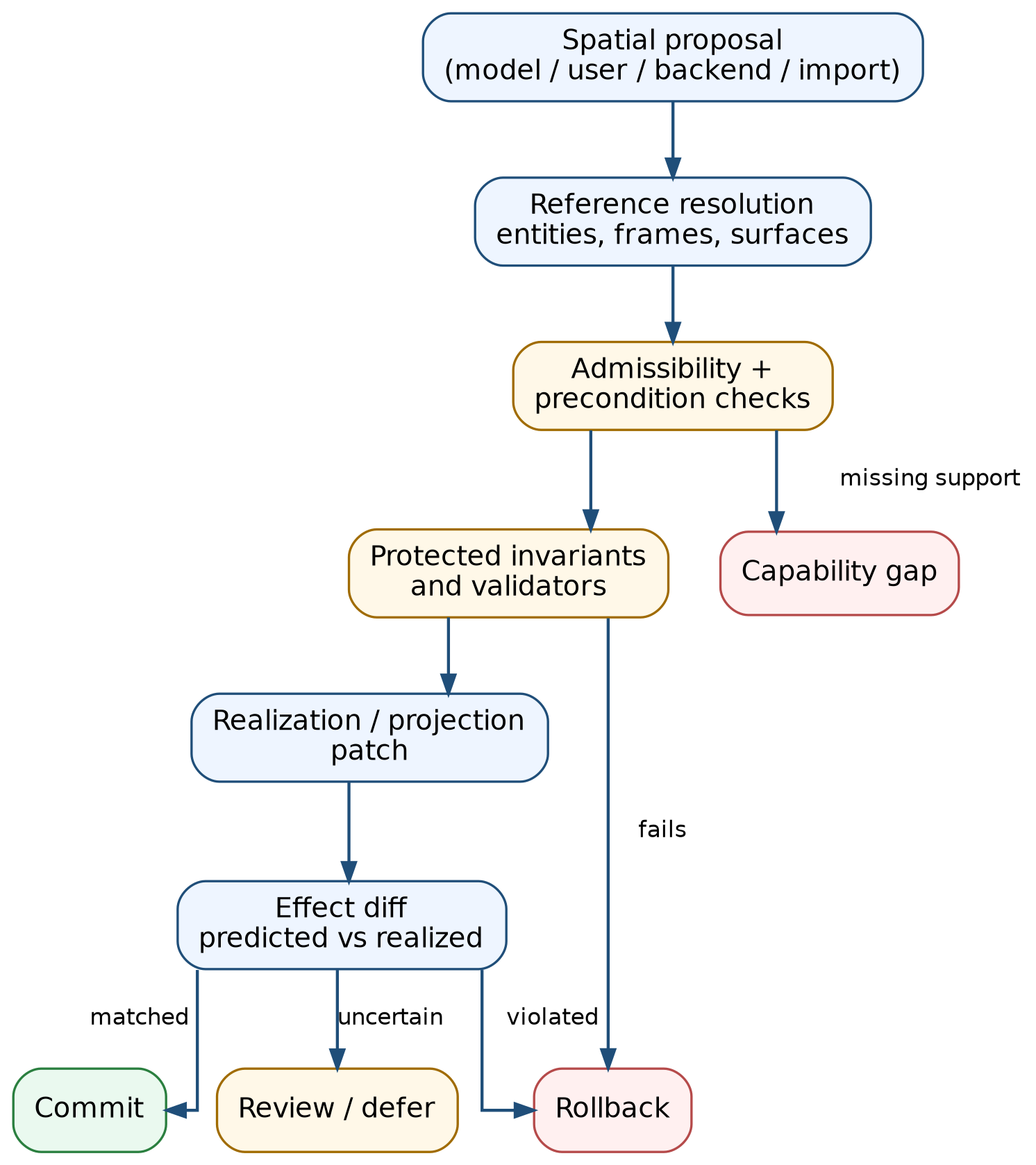}
\caption{Spatial transaction boundary. Proposed changes from models, users, imports, or backends are not treated as spatial truth until they pass reference resolution, admissibility checks, invariant checks, realization/projection, and effect-diff evaluation. The same boundary can commit, defer for review, roll back, or produce a typed capability gap.}
\end{figure}

\subsection{Shared Human And Model
Actuators}\label{shared-human-and-model-actuators}

Hylos avoids separate hidden action spaces for users and models. A human
pointer gesture, a rough drawn region, a model proposal, and a
backend-computed candidate can all enter the same actuator surface.

For example, a generic mark-authoring actuator can represent a path,
region, boundary, point, uncertainty area, blocked zone, access route,
inspection concern, or missing-information marker. The topology of the
mark is distinct from its semantic purpose. Purpose, effect, confidence,
evidence, and unresolved assumptions live in structured qualifiers.

This shared actuator model is important because it prevents a common
failure mode in agentic editors: the user edits shallow viewport state
while the model edits hidden agent state. In Hylos, both are normalized
into draft operations, reviewed or validated, and committed through the
same graph-backed transaction path.

\subsection{Effect Diffs}\label{effect-diffs}

A spatial transaction should not only know that a mutation compiled. It
should know what changed, what was preserved, what unexpectedly changed,
and what could not be checked.

Hylos models this as an effect diff over the transaction:

\begin{verbatim}
predicted effects
actual effects
preserved effects
unexpected effects
unchecked effects
status: matched | review | violated | unchecked
\end{verbatim}

This turns spatial mutation into a state-transition problem rather than
a one-way rendering update. It gives the system a principled way to ask
whether an operation actually did what the model or user expected.

Let \(\hat{\mathcal{F}}\) be the set of predicted effects and
\(\mathcal{F}'\) be the set of observed or audited effects after
realization. Under an effect equivalence relation \(\sim_{\tau}\)
parameterized by tolerance \(\tau\), Hylos computes:

\begin{align*}
\mathcal{F}_{\mathrm{matched}}
  &= \{ f \in \hat{\mathcal{F}} \;|\; \exists f' \in \mathcal{F}' : f \sim_{\tau} f' \}, \\
\mathcal{F}_{\mathrm{unexpected}}
  &= \{ f' \in \mathcal{F}' \;|\; \nexists f \in \hat{\mathcal{F}} : f \sim_{\tau} f' \}, \\
\mathcal{F}_{\mathrm{unchecked}}
  &= \{ f \in \hat{\mathcal{F}} \;|\; \operatorname{Verifier}(f)=\varnothing \}.
\end{align*}

The transaction status is then derived from explicit predicates:

\begin{align*}
\operatorname{matched}(\Delta\mathcal{E})
  &\iff \mathcal{F}_{\mathrm{unexpected}}=\varnothing
      \land \mathcal{F}_{\mathrm{unchecked}}=\varnothing, \\
\operatorname{review}(\Delta\mathcal{E})
  &\iff \mathcal{F}_{\mathrm{unexpected}}\neq\varnothing
      \lor \mathcal{F}_{\mathrm{unchecked}}\neq\varnothing, \\
\operatorname{violated}(\Delta\mathcal{E})
  &\iff \exists i \in \mathcal{I}_{\mathrm{prot}}:
      \neg i(\mathcal{S}_{t+1}), \\
\operatorname{unchecked}(\Delta\mathcal{E})
  &\iff \operatorname{Audit}(\mathcal{M})=\varnothing.
\end{align*}

\section{Evidence-Grounded Interaction
Today}\label{evidence-grounded-interaction-today}

The current prototype implements a scene-scale operability substrate, not
only a local interaction representation. The substrate includes scene
assets, entity hypotheses, surface anchors, spatial assertions, action
candidates, solver jobs, shared actuator invocations, spatial marks,
work artifacts, capability gaps, and effect diffs. Evidence-grounded
interaction is one conservative operating mode over this substrate.

Before any mutation, construction, review item, or deferral is proposed,
the model evaluates candidate interpretations. A candidate describes the
physical interpretation of the task, implicated entities, intended
interaction shape, supporting evidence, risks, uncertainty, and review
triggers. The selected interaction must match the declared interaction
space and be supported by available evidence. If required support is
missing, the correct output is review or deferral rather than invented
geometry.

\subsection{Agentic Evidence
Acquisition}\label{agentic-evidence-acquisition}

In a minimal setting, the model reasons only from existing scene state
and evidence. In an acquisition-enabled setting, the runtime may gather
bounded visual evidence when the current graph suggests that an
interaction is under-constrained and a relevant observation source is
available.

The acquisition pass does not directly change the scene. It records
diagnostic observations, possible implicated entities, plausible causal
paths, and uncertainty. These observations are folded back into
reasoning context only when they connect to declared interactions. This
prevents visual evidence from bypassing the contract layer.

\subsection{Model Spatial Agency
Gates}\label{model-spatial-agency-gates}

Model spatial agency should increase only with available topology and
validation depth:

\textbf{Level 0: Point at or highlight known references.}\\
Commit rule: presentation only.

\textbf{Level 1: Propose marks attached to known references.}\\
Commit rule: review or validator required.

\textbf{Level 2: Request backend-computed geometry.}\\
Commit rule: backend candidate required.

\textbf{Level 3: Propose marks from evidence candidates.}\\
Commit rule: evidence refs plus review required.

\textbf{Level 4: Author richer topology and object-level structure.}\\
Commit rule: active authoring/review path; richer topology must still
pass ingestion, validation, and projection checks.

This gating prevents the model from hallucinating free geometry when the
graph lacks point topology, evidence, anchors, or backend candidates.

\section{Evaluation Method: Repair As Causal Stress
Test}\label{evaluation-method-repair-as-causal-stress-test}

The evaluation uses blind forward replay over a repair task. A canonical
scene is perturbed, the system is asked to repair it, and the resulting
scene is compared against the expected causal and geometric outcome.
This tests whether the agent can reason from evidence and contracts
rather than from hidden task-specific rules.

Repair is emphasized because it creates a falsifiable causal task:

\begin{verbatim}
visible symptom -> upstream cause -> permissible interaction -> validated change
\end{verbatim}

The implemented substrate also supports constructive authoring paths:

\begin{verbatim}
desired outcome -> required topology -> permissible construction -> validated scene state
\end{verbatim}

The key scenario perturbs an existing interaction frame while preserving
the semantic relationship that a dependent component is attached through
that frame. The user instruction is intentionally underspecified:

\begin{verbatim}
The receiving assembly looks laterally wrong relative to the body.
Fix the physical placement.
\end{verbatim}

The scenario is challenging because the visible symptom appears on the
receiving assembly, while the preferred supported repair may live
upstream in the frame that determines its placement.

\subsection{Evaluation Questions}\label{evaluation-questions}

The evaluation is organized around four questions:

\begin{enumerate}
\def\labelenumi{\arabic{enumi}.}
\tightlist
\item
  Can the agent identify that the visible symptom is not necessarily the
  correct edit target?
\item
  Can it select an upstream causal interaction when the scene
  dependencies support that interpretation?
\item
  Does validation prevent unsupported geometry changes and force
  deferral when support is missing?
\item
  Can a new generic spatial alternative resolve an ambiguity without
  becoming a product-specific rule?
\end{enumerate}

\subsection{Baselines And Conditions}\label{baselines-and-conditions}

The public evaluation is organized around conceptual controls rather
than a large benchmark suite. These controls isolate the design choices
needed for the causal repair proof and define the comparison structure
for a broader benchmark over the existing substrate.

\textbf{Condition: Direct geometry edit.}\\
Description: the agent edits the visible misplaced component directly.\\
Expected failure mode: may improve appearance while missing the upstream
cause.

\textbf{Condition: Prompt heuristic.}\\
Description: the agent receives task-specific placement guidance in
prose.\\
Expected failure mode: can pass the scene while hiding the reasoning
path.

\textbf{Condition: Scene structure only.}\\
Description: the agent sees structured state but lacks enforced
interaction constraints.\\
Expected failure mode: may identify the cause but still propose
unsupported edits.

\textbf{Condition: Contract-bounded interaction.}\\
Description: the agent must act through permissible interactions and
validation.\\
Expected failure mode: defers when support is missing.

\textbf{Condition: Contract-bounded interaction with evidence
acquisition.}\\
Description: the agent may request bounded diagnostic evidence before
selecting an interaction.\\
Expected failure mode: can still defer if the new evidence does not
connect to a permissible interaction.

\textbf{Condition: Contract-bounded interaction with reusable spatial
alternatives.}\\
Description: the agent can choose among generic supported geometric
alternatives.\\
Expected failure mode: should resolve valid ambiguity without
product-specific rules.

These conditions are used as an interpretive framework for the public
artifact study. The paper reports qualitative evidence for the causal
repair trajectory, while the broader internal fixture suite motivates
benchmark packaging across object classes, relation structures, mutation
families, authoring tasks, and generated-asset regimes.

\subsection{Evaluation Criteria}\label{evaluation-criteria}

A replay is considered successful only if it satisfies all of the
following criteria:

\begin{itemize}
\tightlist
\item
  The selected interpretation identifies the upstream driver rather than
  only the visible symptom.
\item
  The chosen interaction is inside the declared interaction space.
\item
  The chosen geometric alternative is supported by scene evidence or
  scene structure.
\item
  Unsupported alternatives are rejected or deferred rather than applied.
\item
  The realized scene passes manual visual inspection for the intended
  spatial interpretation.
\end{itemize}

\subsection{Experimental Trajectory}\label{experimental-trajectory}

The experiment was developed as a sequence of increasingly constrained
replays rather than as a single demonstration. Each replay removed an
informal source of help or introduced a more explicit contract boundary.
The goal was to test whether the model could still identify the causal
interaction without relying on product-specific hints.

The first phase established the task: a dependent component was visibly
misplaced, and the system had to decide whether the visible component
itself should move or whether a supporting upstream relation should
change. This phase tested the basic relational hypothesis: spatial
repair should be routed through causal scene structure, not through the
nearest visible geometry.

The second phase removed prompt-side placement heuristics and required
the model to reason from structured state, evidence, permissible
interactions, and validation constraints. Unsupported numeric edits were
rejected or deferred. This was important because a plausible-looking
edit is not a successful repair if it bypasses the scene contract.

The third phase introduced bounded evidence acquisition. When existing
context was insufficient, the system could acquire a diagnostic
observation, but that observation still had to connect back to
permissible interactions. This tested whether additional evidence could
improve spatial reasoning without giving the model a free path to
arbitrary geometry edits.

The fourth phase came from manual visual review of a successful replay.
The system had selected a supported alignment, but the visible goal
exposed a second legitimate interpretation: centering relative to the
owning body rather than aligning to an external reference. The system
was then extended with a generic geometry-derived alternative. This did
not add a product rule; it added a reusable spatial option that the
model could select and the validator could check.

\section{Result: Causal Repair Through The Operability
Contract}\label{result-causal-repair-through-the-operability-contract}

The successful replay followed this causal chain:

\begin{verbatim}
visual observation
-> diagnostic evidence for lateral placement mismatch
-> dependency structure identifies an upstream placement driver
-> declared interaction space permits changing that driver
-> supported geometric alternative is selected
-> validation accepts the interaction
\end{verbatim}

The important result is not the particular object being repaired. It is
that the model did not directly move the visible dependent component. It
traced the visible symptom through semantic dependency structure and
selected a supported upstream change.

\begin{figure}[tbp]
\centering
\includegraphics[width=0.92\linewidth]{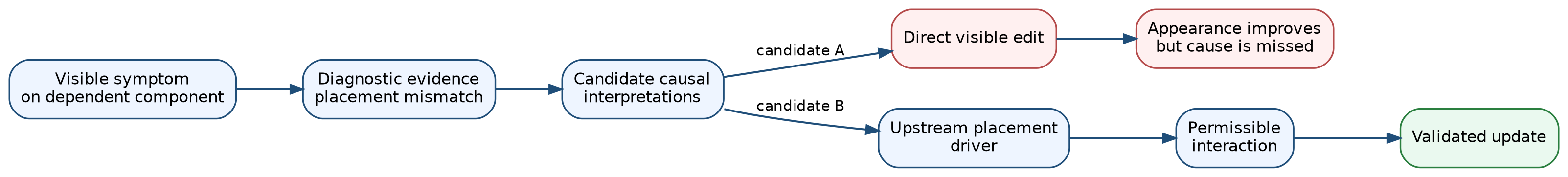}
\caption{Causal repair chain. The key empirical stress test is not whether a visible component can be moved, but whether the agent routes a visible symptom through candidate causal interpretations to the supported upstream driver and rejects weaker local edits.}
\end{figure}

\subsection{Qualitative Results}\label{qualitative-results}

The artifact study is summarized as a claim-evidence matrix rather than
a statistical benchmark. The matrix is written as cards instead of a
markdown pipe table because long technical cells render poorly in some
preview panes.

\textbf{Claim 1: The visible symptom need not be the correct edit
target.}\\
Evidence: the successful replay routes the repair through an upstream
placement driver rather than directly moving the dependent component.\\
Public benchmark status: this paper uses one focused repair family as a
clear causal artifact while the broader substrate supports additional
mutation and authoring scenarios.

\textbf{Claim 2: Contract boundaries prevent unsupported geometry
changes.}\\
Evidence: intermediate runs deferred or rejected unsupported
alternatives instead of applying arbitrary numeric edits.\\
Public benchmark status: the mechanism is exercised in prototype
fixtures; the next publication target is repeated reporting across more
object and relation types.

\textbf{Claim 3: Additional evidence helps only when it reconnects to
permissible interactions.}\\
Evidence: diagnostic evidence is useful when it supports the same causal
path as the declared interaction space.\\
Public benchmark status: evidence acquisition is present in the runtime
and should be reported across richer ambiguity classes in the public
benchmark suite.

\textbf{Claim 4: Generic spatial alternatives can resolve ambiguity
without product rules.}\\
Evidence: manual review exposed a reference-alignment vs body-centering
ambiguity; a reusable geometry-derived alternative resolved the visible
goal.\\
Public benchmark status: this is an expandable alternative library
rather than a product-specific rule path.

\textbf{Claim 5: The same abstraction supports repair, authoring,
mutation, and variant generation.}\\
Evidence: repair exercises symptom-to-cause reasoning, while internal
fixtures exercise intent-to-topology reasoning, direct mutation, frame
transforms, support-region changes, multi-region consequence reasoning,
and variant-generation scenarios over the same public abstraction.\\
Public benchmark status: this paper reports the causal repair trajectory
most deeply; the broader fixture suite should be packaged as the next
public benchmark release.

This matrix separates the reported public artifact from the broader
implemented substrate and from the benchmark breadth that should be
published next.

\subsection{What Failed Usefully}\label{what-failed-usefully}

Several intermediate outcomes were informative. When the system lacked a
supported geometric alternative for the selected interpretation,
deferral was the correct result. When an apparent repair targeted the
wrong level of the scene, validation prevented that interpretation from
becoming an arbitrary edit. These failures were not incidental; they
were evidence that the contract boundary was doing useful work.

This is why the evaluation emphasizes blind forward replay. The agent is
not given a hidden answer key or a hand-authored instruction to choose
the upstream driver. It must infer the relevant physical interpretation
from exposed scene structure and evidence, then act only through an
admissible interaction.

\subsection{What The Experiment Shows}\label{what-the-experiment-shows}

The final result supports three claims. First, a foundation model can
use semantic scene dependencies to reason about causes that are not
located on the visible symptom. Second, validation can keep the model
inside a declared interaction space without reducing it to a scripted
rule engine. Third, new reusable spatial alternatives can be added to
the deterministic layer without changing the task into a
product-specific heuristic.

The public artifact study is deliberately scoped around one causal
repair family. Its value is that it makes the abstraction observable
while connecting to a broader implemented substrate that already
exercises authoring, mutation, recovery, projection, and
consequence-aware interaction paths.

\section{From Wrapped Neural Assets To Model-Native Spatial
Artifacts}\label{from-wrapped-neural-assets-to-model-native-spatial-artifacts}

The current transaction architecture is not the end state. It is a
reliability scaffold for the transition from explicit graph operations
to future model-native spatial artifacts.

\subsection{Stage 1: Transaction-Safe Explicit
Lowering}\label{stage-1-transaction-safe-explicit-lowering}

In the current regime, the model selects or proposes bounded graph
operations. The runtime validates the operation, resolves references,
derives a compatible graph mutation or realization patch, and commits
only if the transaction passes its checks.

This stage is conservative but powerful. It prevents the model from
becoming an unbounded script generator and creates a stable audit trail
for every geometry-changing action.

\subsection{Stage 2: Structure Recovery Over Neural
Assets}\label{stage-2-structure-recovery-over-neural-assets}

Generated meshes, splats, neural fields, and imported scans can enter
the system as realization assets, but they should not be treated as
complete scene truth. Hylos wraps them with recovered structure:

\begin{itemize}
\tightlist
\item
  functional axes and grounding surfaces;
\item
  contact and support frames;
\item
  uncertainty regions;
\item
  candidate entities and boundaries;
\item
  access, obstruction, or interaction regions;
\item
  provenance and acquisition status.
\end{itemize}

The result is a hybrid realization asset: visually rich geometry plus a
deterministic operability shell. Unsupported or ambiguous claims remain
hypotheses or capability gaps rather than fake geometry.

\subsection{Stage 3: Hybrid
Co-Generation}\label{stage-3-hybrid-co-generation}

In the hybrid regime, models begin to co-generate visual geometry and
symbolic operability structure. A generated scene may include a mesh or
neural field alongside entities, frames, constraints, paths, clearances,
and candidate edit handles.

The key requirement is coincidence: symbolic claims must map to the
geometry they describe, and the runtime must be able to validate or
reject those mappings. Co-generation is useful only if the artifact can
still answer operational questions about evidence, validity,
uncertainty, and effects.

\subsection{Stage 4: Model-Native Spatial
Artifacts}\label{stage-4-model-native-spatial-artifacts}

The future target is a model-native spatial artifact:

\begin{verbatim}
GeneratedSpatialArtifact = {
  geometry,
  topology,
  operational constraints,
  editable handles,
  provenance and uncertainty,
  audit hooks
}
\end{verbatim}

Such an artifact would not be a raw mesh with labels attached afterward.
It would be generated as an operable object or world: geometry, internal
structure, constraints, and edit handles produced together.

This is a model-native target contract rather than a claim of fully
solved model capability. The Hylos operability contract defines what
such artifacts must provide as spatial generation moves from visual
plausibility toward reliable object, assembly, and environment
operation.

\begin{figure}[tbp]
\centering
\includegraphics[width=0.74\linewidth]{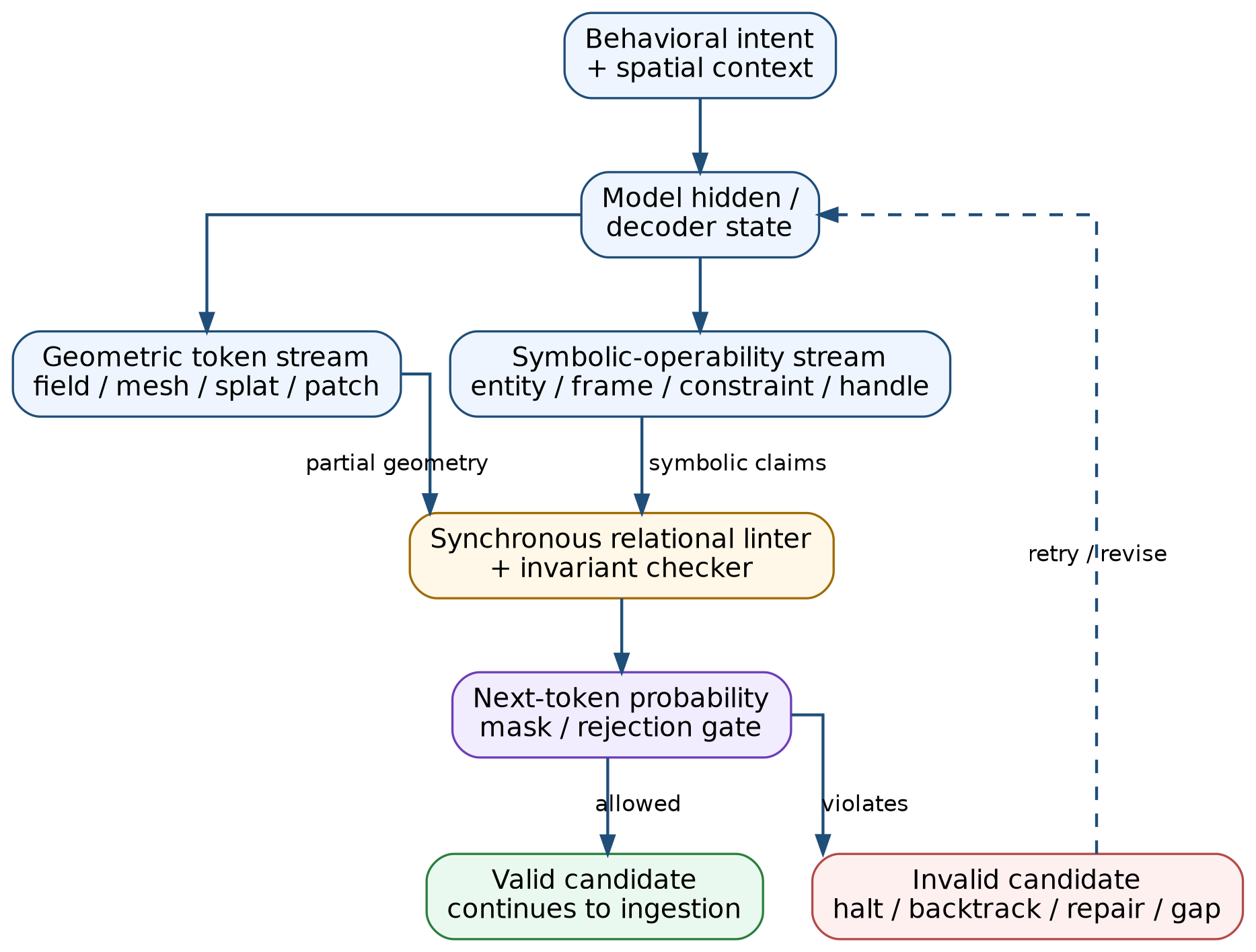}
\caption{Dual-stream token-level spatial validity guardrail. A model-native spatial generator can emit geometric and symbolic-operability streams while a synchronous linter or invariant checker masks invalid next-token candidates before broken spatial artifacts are rendered, exported, simulated, or committed.}
\end{figure}

More formally, let a generated spatial artifact be:

\[\mathcal{A}_{spatial} = \left(\mathbf{G}, \mathbf{S}, \mathbf{C}, \mathbf{H}, \mathbf{P}, \mathbf{U}\right)\]

where \(\mathbf{G}\) is the geometric realization field, \(\mathbf{S}\)
is the symbolic topology and component graph, \(\mathbf{C}\) is the
operational constraint set, \(\mathbf{H}\) is the editable handle set,
\(\mathbf{P}\) is provenance, and \(\mathbf{U}\) is uncertainty. The
artifact is operable only when there exists a runtime ingestion map:

\[\Phi_{ingest}: \mathcal{A}_{spatial} \rightarrow \mathcal{S}_{t+1} \cup \{\text{review}, \text{capability gap}\}\]

such that geometry, symbolic claims, constraints, and handles can be
grounded into the scene operability graph with explicit uncertainty and
evidence status.

\subsection{Token-Level Neurosymbolic
Guardrails}\label{token-level-neurosymbolic-guardrails}

In a model-native architecture, spatial validity should be checked
before a full artifact is rendered, exported, or physically acted upon.
Hylos therefore treats future spatial generation as coupled decoding
over geometric and symbolic token streams:

\[z_k = (t^g_k, t^s_k)\]

where \(t^g_k\) is a geometric token and \(t^s_k\) is a
symbolic-operability token. The geometric stream may encode surface
fields, occupancy, splats, mesh patches, or coordinate quantization. The
symbolic stream encodes entities, frames, contact claims, clearances,
constraints, edit handles, and provenance links.

Let the unconstrained model distribution be:

\[p_0(z_{k+1} \mid z_{\leq k}) = \operatorname{softmax}(W h_k)\]

Let \(\mathcal{L}_{engine}\) be a relational logic engine and
\(\mathcal{I}_{global}\) be the active invariant set. A token candidate
receives a validity mask:

\[m_k(z) = \operatorname{Ind}\!\left(\mathcal{L}_{engine}(\operatorname{Sym}(z_{\leq k} \cup \{z\})) \vdash \mathcal{I}_{global}\right)\]

The Hylos-constrained next-token distribution is:

\[
 p_H(z_{k+1}=z \mid z_{\leq k}) =
 \frac{p_0(z \mid z_{\leq k})\,m_k(z)}{\sum_{z' \in \mathcal{V}} p_0(z' \mid z_{\leq k})\,m_k(z')}.
\]

If the denominator is zero, the runtime has reached a dead-end under
current invariants. The correct outcome is not to emit invalid geometry.
It is to halt, backtrack, request review, or record a typed capability
gap.

\subsection{Latent Execution Auditing And Test-Time
Repair}\label{latent-execution-auditing-and-test-time-repair}

Some failures are not best handled by hard masking alone. A candidate
artifact may be close to valid but violate a differentiable clearance,
support, or containment objective. In that case, Hylos can formulate
test-time repair as constrained latent optimization.

Let \(\theta\) be model parameters, \(z\) be the latent state,
\(\mathcal{D}_{\theta}(z)\) decode a candidate artifact, and
\(v_i(\mathcal{A})\) be violation functions over decoded geometry and
symbolic claims. A local repair step solves:

\[
\begin{aligned}
z^* = \arg\min_{z'} \;& \mathcal{L}_{task}(\mathcal{D}_{\theta}(z'), y) \\
&+ \lambda \sum_i \max(0, v_i(\mathcal{D}_{\theta}(z')))^2 \\
&+ \beta \lVert z' - z \rVert_2^2.
\end{aligned}
\]

where \(\mathcal{L}_{task}\) preserves the intended artifact, the
violation penalty pushes the artifact back into the valid region, and
the proximity term prevents unrelated regeneration. The repaired
artifact still must pass runtime ingestion and transaction validation
before commit.

\begin{figure}[tbp]
\centering
\includegraphics[width=0.70\linewidth]{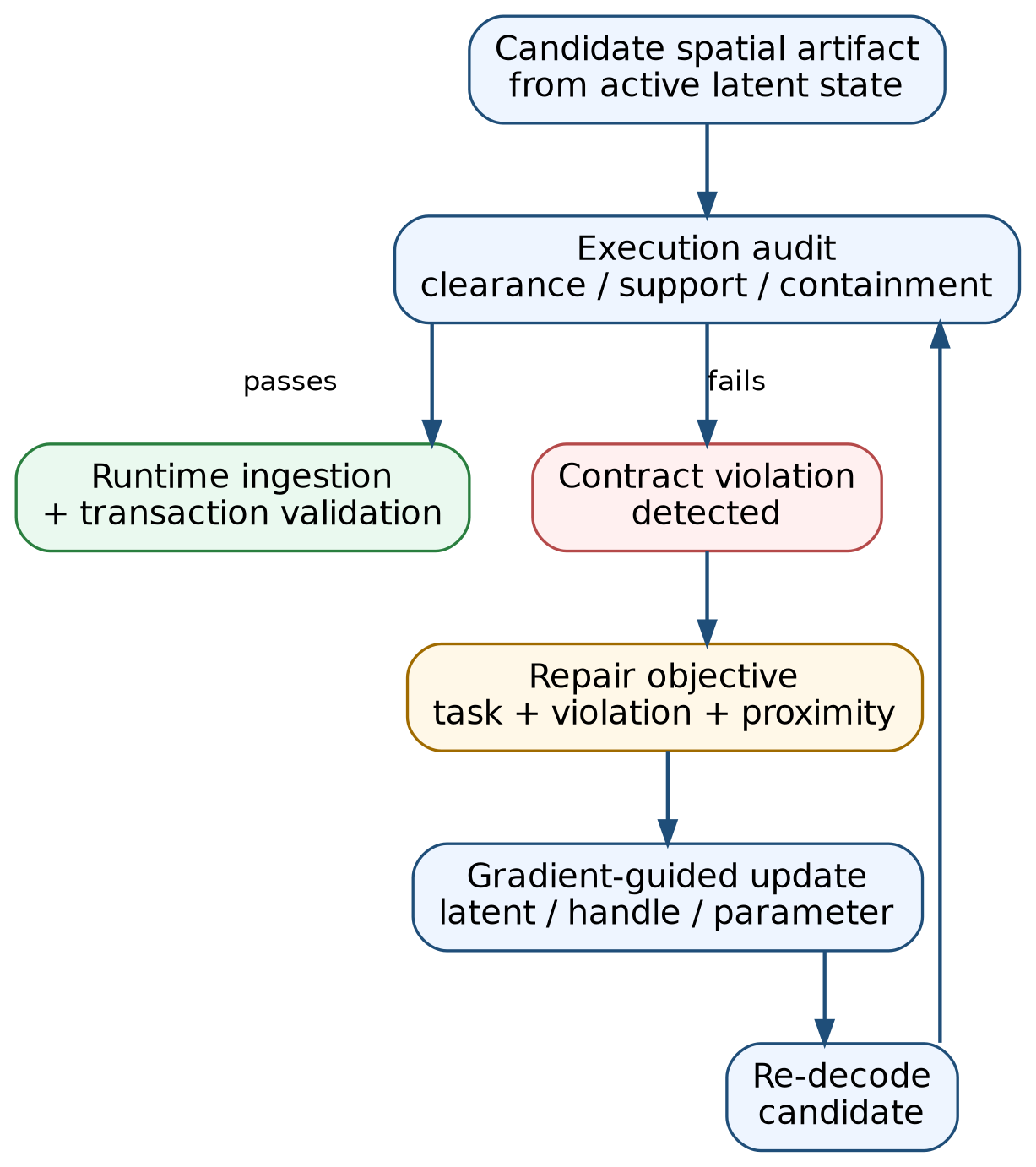}
\caption{Latent execution auditing and test-time repair. Candidate artifacts that fail execution audits can be repaired by optimizing against task preservation, violation penalties, and proximity constraints, then re-decoded and re-audited before runtime ingestion.}
\end{figure}

\subsection{Parametric Neural
Jacobians}\label{parametric-neural-jacobians}

Classical parametric CAD exposes edit variables that rebuild a feature
tree. Model-native geometry lacks such an explicit feature tree unless
editability is generated as part of the artifact. Hylos therefore treats
editable handles as first-class outputs.

Let \(\mathbf{x}(u; z) \in \mathbb{R}^3\) be a decoded point on the
generated spatial field at local coordinate \(u\), and let
\(\mathbf{h}_j(z) \in \mathbb{R}^d\) be an editable handle. A local edit
\(\Delta \mathbf{h}_j\) induces a first-order deformation:

\[\Delta \mathbf{x}(u) \approx J_j(u; z) \Delta \mathbf{h}_j\]

where:

\[J_j(u; z) = \frac{\partial \mathbf{x}(u; z)}{\partial \mathbf{h}_j}\]

The handle is valid only if the resulting deformation preserves required
constraints:

\[\operatorname{HandleValid}(j) \iff \forall c \in \mathbf{C}_{local}, \quad c\left(\mathbf{x}(u; z) + J_j(u; z)\Delta\mathbf{h}_j\right) \leq 0\]

This defines the contract model-native spatial artifacts should satisfy:
edit handles expose predictable local deformation fields and are audited
against symbolic constraints before commit.

\begin{figure}[tbp]
\centering
\includegraphics[width=0.70\linewidth]{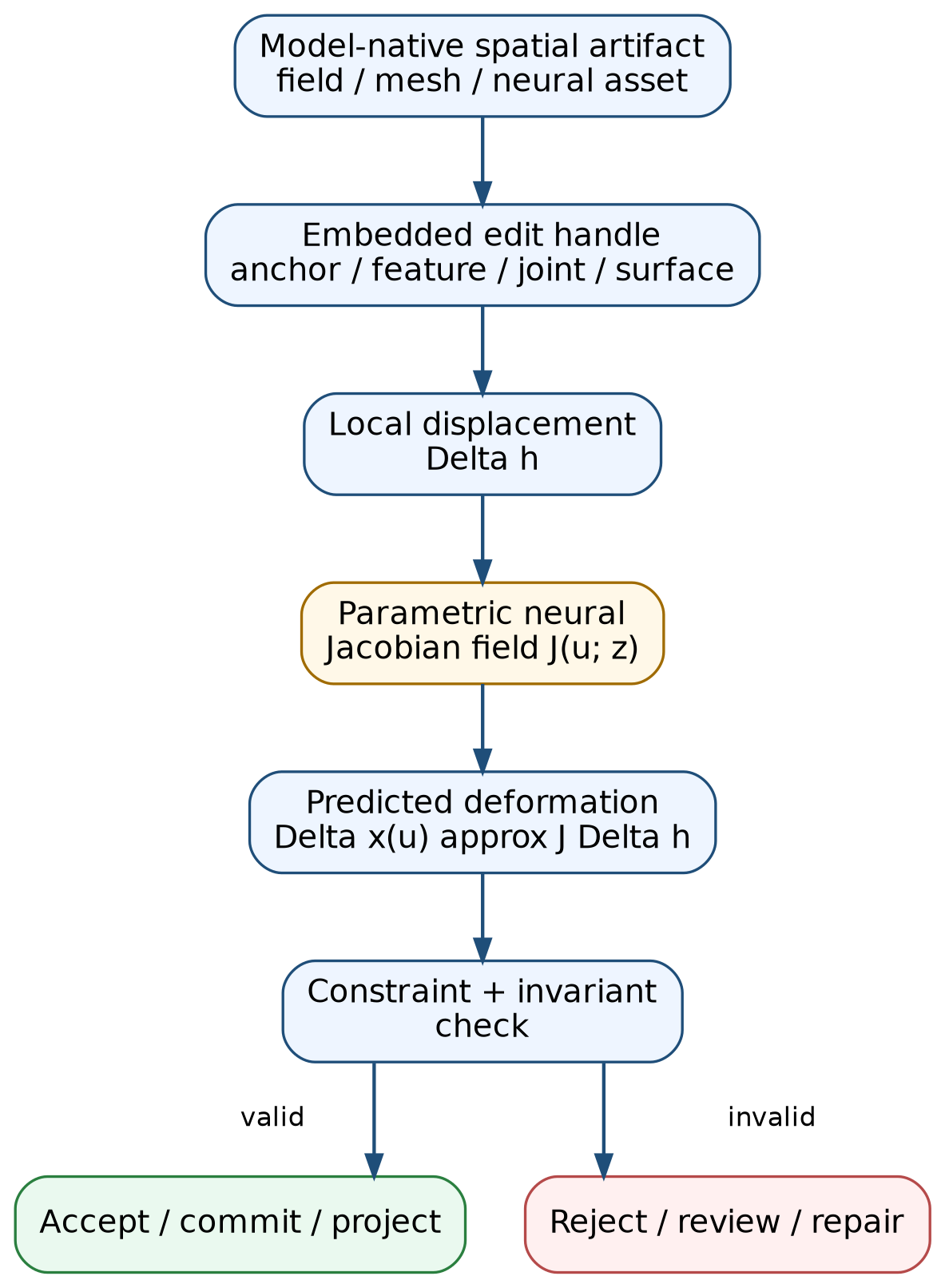}
\caption{Parametric neural edit handles. Instead of relying only on classical CAD feature trees, a model-native artifact may expose embedded handles whose local deformations are predicted through a neural Jacobian or equivalent sensitivity field and checked against constraints before commit.}
\end{figure}

\subsection{Self-Supervised Cycle
Consistency}\label{self-supervised-cycle-consistency}

The model-native regime also requires training signals that are not
limited to hand-authored compiler outputs. Hylos frames this as cycle
consistency between generation, simulation, structure recovery, and
re-ingestion.

Let \(\mathcal{A} = \mathcal{D}_{\theta}(z)\) be a generated artifact,
\(\Pi_j\) be a projection adapter into a target environment such as
rendering, CAD, robotics simulation, or physics simulation, and \(\Psi\)
be the recovery/ingestion operator back into operability state. A
cycle-consistency objective can be written:

\[\mathcal{L}_{cycle} = \sum_j d\left(\Psi(\Pi_j(\mathcal{A})), \operatorname{Op}(\mathcal{A})\right)\]

where \(\operatorname{Op}(\mathcal{A})\) extracts the intended
operability structure from the artifact, and \(d\) measures disagreement
in recovered entities, frames, constraints, handles, and effects.
Simulation rewards or penalties can be added:

\[\mathcal{L}_{train} = \mathcal{L}_{gen} + \alpha \mathcal{L}_{cycle} + \gamma \mathcal{L}_{sim} + \eta \mathcal{L}_{constraint}\]

This creates a research path by which generated spatial artifacts can be
trained not only to look plausible, but to remain recoverable,
projectable, editable, and validatable.

\begin{figure}[tbp]
\centering
\includegraphics[width=0.92\linewidth]{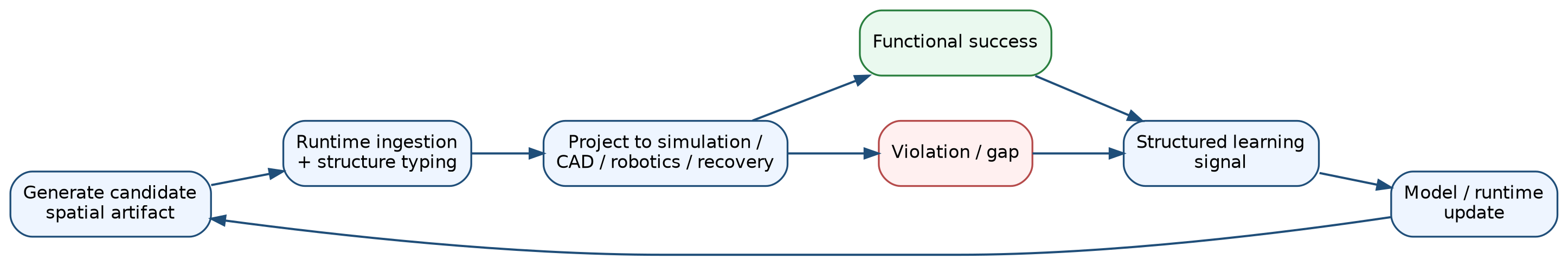}
\caption{Self-supervised operability loop. Generated artifacts are ingested, projected into downstream environments, scored as functional successes or violations, and converted into structured signals that improve future model behavior and runtime coverage.}
\end{figure}

\subsection{Runtime Ingestion Beats Raw
Generation}\label{runtime-ingestion-beats-raw-generation}

Even model-native artifacts should not become truth directly. They must
be ingested into the runtime, checked against global invariants,
assigned uncertainty and provenance, projected through adapters, and
committed or rejected.

This reframes the relationship between future spatial models and
deterministic systems. The goal is not to remove validation. The goal is
to make generated artifacts more natively validatable.

\begin{verbatim}
model-native artifact
-> runtime ingestion
-> structure, uncertainty, and provenance typing
-> invariant validation
-> effect diff
-> committed scene truth or review/capability gap
\end{verbatim}

\section{Scientific Evaluation
Program}\label{scientific-evaluation-program}

The causal repair study is a minimal public empirical anchor, not a
complete validation program. The current prototype already exercises
more than the repair family through internal fixtures for mutation,
frame transforms, support-region changes, multi-region consequence
reasoning, and variant generation. The evaluation program should
therefore be understood as a packaging and scaling task: converting the
implemented substrate into repeatable public benchmarks across object
classes, relation structures, asset types, transaction families, and
model regimes.

\subsection{Repair Benchmark}\label{repair-benchmark}

Repair tasks test whether the model can route visible symptoms through
causal scene dependencies:

\begin{verbatim}
observed issue on object A
-> A depends on B
-> B is driven by frame or constraint C
-> C has a supported interaction
-> interact with C, not A
\end{verbatim}

\subsection{Authoring Benchmark}\label{authoring-benchmark}

Authoring tasks test whether desired outcomes become required topology,
surfaces, openings, attachments, constraints, and validation checks:

\begin{verbatim}
desired outcome for object A
-> A requires relation to B
-> B requires frame, boundary, or affordance C
-> C has a supported construction or mutation
-> create or mutate C, then validate A's behavior
\end{verbatim}

\subsection{Structure Recovery
Benchmark}\label{structure-recovery-benchmark}

Structure recovery tasks should evaluate whether unstructured generated
or imported assets can be wrapped with useful operability structure:

\begin{itemize}
\tightlist
\item
  recovered support/contact frames;
\item
  entity and boundary candidates;
\item
  uncertainty regions;
\item
  access and obstruction fields;
\item
  provenance links;
\item
  correct deferrals when structure cannot be recovered.
\end{itemize}

\subsection{Model-Native Artifact Ingestion
Benchmark}\label{model-native-artifact-ingestion-benchmark}

Future model-native artifacts should be evaluated by more than visual
quality. Candidate metrics include:

\begin{itemize}
\tightlist
\item
  geometric-symbolic consistency;
\item
  edit-handle validity;
\item
  constraint satisfaction;
\item
  projection fidelity across display, CAD, simulation, and robotics
  views;
\item
  effect-diff accuracy;
\item
  rate of unsupported claims converted into explicit capability gaps
  rather than fake geometry.
\end{itemize}

\subsection{Quantitative Metrics}\label{quantitative-metrics}

The evaluation program should report metrics that correspond to
operability, not only perceptual quality.

For causal repair, let \(a_i^*\) be the expected upstream admissible
actuator for scenario \(i\), \(\hat{a}_i\) be the selected actuator, and
\(\operatorname{Valid}(\hat{a}_i)\) indicate that the transaction passed
validation:

\[\operatorname{CRA} = \frac{1}{N} \sum_{i=1}^{N} \operatorname{Ind}\left(\hat{a}_i = a_i^* \land \operatorname{Valid}(\hat{a}_i)\right)\]

where \(\operatorname{CRA}\) is causal repair accuracy. Because good
deferral is also a success mode, let \(u_i\) indicate that scenario
\(i\) is under-supported and should not commit. Deferral precision is:

\[\operatorname{DP} = \frac{\sum_i \operatorname{Ind}(\hat{d}_i = 1 \land u_i = 1)}{\sum_i \operatorname{Ind}(\hat{d}_i = 1)}\]

For generated or recovered artifacts, geometric-symbolic consistency
measures whether symbolic claims correspond to realized geometry:

\[\chi_{gsc} = \frac{1}{|\mathbf{C}|} \sum_{c \in \mathbf{C}} \operatorname{Ind}\left(\operatorname{CheckGeom}(c, \mathbf{G}) \leq \epsilon_c\right)\]

Handle invariant preservation measures whether edit handles preserve
local constraints under sampled edits:

\[\Gamma_{hip} = \frac{1}{|\mathbf{H}| |\Delta|} \sum_{h \in \mathbf{H}} \sum_{\delta \in \Delta} \operatorname{Ind}\left(\operatorname{Preserve}(\mathbf{C}_{local}, h, \delta)\right)\]

Effect-diff precision and recall compare predicted effects
\(\hat{\mathcal{F}}\) and audited effects \(\mathcal{F}'\):

\[
\begin{aligned}
\operatorname{Prec}_{effect} &=
\frac{|\mathcal{F}_{matched}|}{|\mathcal{F}_{matched}| + |\mathcal{F}_{unexpected}|},\\
\operatorname{Rec}_{effect} &=
\frac{|\mathcal{F}_{matched}|}{|\mathcal{F}_{matched}| + |\mathcal{F}_{missed}|}.
\end{aligned}
\]

Finally, transaction commit success should be reported separately from
raw generation success. Let \(\mathcal{A}_{commit}\) be the submitted
artifacts that pass ingestion and commit, and let
\(\mathcal{A}_{submitted}\) be the full submitted set:

\[\operatorname{TCS} = \frac{|\mathcal{A}_{commit}|}{|\mathcal{A}_{submitted}|}\]

This prevents a model from receiving credit for visually plausible
outputs that cannot become validated scene truth.

\subsection{Ablations}\label{ablations}

The evaluation program should compare:

\begin{itemize}
\tightlist
\item
  prompt-only spatial editing;
\item
  direct script generation;
\item
  direct mesh or neural asset generation;
\item
  scene structure without enforced transactions;
\item
  transactions without evidence acquisition;
\item
  full contract-bounded interaction;
\item
  model-native artifacts with and without runtime ingestion.
\end{itemize}

\section{Discussion}\label{discussion}

\subsection{Architecture Is Not The Final
Product}\label{architecture-is-not-the-final-product}

The current Hylos runtime is already a working substrate for reliable
spatial interaction. The larger thesis is broader: spatial intelligence
should produce operable artifacts by default. The transaction layer is
therefore not a retreat from model-native generation. It is the
reliability boundary that defines what model-native generation must
satisfy as it scales from objects and assemblies to scenes and worlds.

\subsection{Why This Is Not A Product-Specific
Rule}\label{why-this-is-not-a-product-specific-rule}

The evaluated behavior does not rely on a hardcoded product rule such as
``center this particular component.'' The deterministic system exposes
scene relationships, permissible interactions, provenance, and supported
alternatives. The model selects an interpretation. The validation layer
accepts only interactions consistent with the structured scene contract.

\subsection{Good Deferral Remains
Important}\label{good-deferral-remains-important}

Earlier runs correctly deferred when the required supported alternative
was missing. This is desirable. The system should not force a geometry
change merely because a visual issue exists. It should act only when the
scene contract and evidence support the selected interaction.

For future model-native systems, this principle becomes even more
important. A generated artifact that cannot support its own claims
should not be silently accepted. It should be partially ingested, marked
uncertain, routed for review, or rejected with a typed capability gap.

\section{Current Boundaries And Evaluation
Scope}\label{current-boundaries-and-evaluation-scope}

This work is an architecture and artifact-study contribution built on an
implemented prototype substrate. The main boundary is not the absence of
scene-scale operability machinery; it is the current public packaging
and benchmark breadth. The paper reports a focused causal repair
artifact because it makes the abstraction observable. The prototype
itself already contains broader scene-scale state, action search,
evidence, review, recovery, projection, and consequence-tracking
mechanisms.

Current boundaries:

\begin{itemize}
\tightlist
\item
  The public empirical anchor emphasizes one causal repair family. The
  broader prototype includes fixtures for direct mutation, frame
  transforms, support-region changes, multi-region consequence
  reasoning, constructive authoring, placement, and variant generation.
  These should be packaged into a broader public benchmark rather than
  treated as absent capabilities.
\item
  The current prototype includes scene-scale operability state and an
  action search space: assets, entity hypotheses, anchors, assertions,
  action candidates, solver jobs, shared actuator invocations, marks,
  work artifacts, effect diffs, and review/capability states. The
  remaining work is to standardize coverage across arbitrary object
  classes, relation families, validators, and backend realization paths.
\item
  Structure recovery and evidence acquisition are already present for
  imported meshes, splats, collider-backed substrates, mesh-hit
  clusters, support evidence, visible candidates, and action candidates.
  The open research target is broad evaluation across more neural asset
  classes, imported scene sources, noisy scans, and ambiguity regimes.
\item
  Relation, assertion, and action structures are present in the runtime.
  The evaluation suite should expand their public coverage across
  containment, adjacency, attachment, articulation, clearance, flow,
  actuation, causal dependency, object-level part relationships, and
  environment-level interaction paths.
\item
  Constructive authoring, placement, mutation, resizing, and
  variant-generation scenarios exist in the system. The public paper
  reports the causal repair trajectory most deeply, while the next
  benchmark should report intent-to-topology conversion across surfaces,
  openings, attachments, constraints, object features, assembly
  relations, and realization outcomes.
\item
  Review, deferral, unresolved assertions, solver status, and capability
  gaps are first-class runtime concepts. The user-facing product
  workflow and public benchmark scoring protocol for repeated review,
  correction, approval, and multi-step recovery are still maturing.
\item
  Constraint solving, contact reasoning, clearance reasoning, collision
  reasoning, and consequence checks are implemented for supported cases.
  They are not yet complete across every class of physical scene, object
  assembly, manufacturing constraint, and object-environment
  interaction.
\item
  Backend and projection coverage is expanding. A valid spatial
  operation may still encounter missing lowerers, unsupported backend
  operations, projection gaps, or reference-fidelity gaps after
  realization.
\item
  Visual diagnostics may miss evidence in complex scenes. Diagnostic
  observations are useful only when they reconnect to admissible
  interactions, evidence references, and validation paths.
\item
  Model-native spatial artifacts are proposed as a target contract, not
  claimed as a solved foundation-model capability.
\item
  The system depends on the correctness of its scene representation,
  validation layer, evidence acquisition, compiler/lowering process,
  backend realization, and audit procedure.
\item
  The final output is a structured scene mutation or artifact ingestion
  result, not full physical certification.
\end{itemize}

\section{Scaling And Public Evaluation
Roadmap}\label{scaling-and-public-evaluation-roadmap}

The next research step is to turn the existing Hylos substrate into a
broad public evaluation program for operable physical 3D. The emphasis
is scale, formalization, standardization, benchmark release, and
cross-domain coverage. Hylos already exercises the core pattern:
scene-scale operability state, action candidates, solver jobs, shared
actuator invocations, structure recovery, review/deferral states, effect
diffs, and projection paths. The roadmap is to make this breadth
measurable, repeatable, and comparable.

\begin{enumerate}
\def\labelenumi{\arabic{enumi}.}
\tightlist
\item
  \textbf{Relation graph coverage at object and environment scale:}
  expand reporting over existing relation and assertion structures to
  cover containment, adjacency, attachment, articulation, support,
  clearance, flow, actuation, part-level dependencies, assembly
  constraints, and environment-level causal links.
\item
  \textbf{Constructive authoring benchmarks:} package existing
  authoring, placement, mutation, resizing, and variant-generation
  scenarios into public tests for intent-to-topology conversion across
  surfaces, openings, attachments, object features, constraints, and
  realization outcomes.
\item
  \textbf{Causal and goal graph evaluations:} report how the
  assertion/action/solver substrate links observed issues and desired
  outcomes to plausible drivers, requirements, validators, and
  admissible interactions across repair, authoring, inspection,
  optimization, and routing tasks.
\item
  \textbf{Transaction graph standardization:} formalize the current
  action and transaction substrate into standardized preconditions,
  protected invariants, effect assertions, rollback semantics, audit
  records, and backend realization contracts across object-level and
  scene-level operations.
\item
  \textbf{Evidence acquisition benchmarks:} measure when bounded visual
  or geometric evidence improves spatial reasoning under controlled
  ambiguity, and when the correct behavior is review, deferral, or
  additional acquisition.
\item
  \textbf{Uncertainty, review, and capability-gap reporting:}
  standardize runtime uncertainty, review, deferral, unresolved
  assertions, solver status, and capability-gap outputs so they can be
  scored consistently across transaction families.
\item
  \textbf{Cross-representation adapter suites:} extend existing
  realization and preview projection paths into a formal suite for
  display, CAD/export, simulation, robotics, manufacturing, inspection,
  sales visualization, training environments, and audit views.
\item
  \textbf{Structure recovery benchmarks:} evaluate recovery over
  imported meshes, splats, scans, collider-backed substrates, generated
  assets, and neural representations, measuring whether recovered
  entities, frames, surfaces, relationships, uncertainty states, and
  action candidates support downstream operation.
\item
  \textbf{Model-native artifact contracts:} define output formats,
  training objectives, and ingestion checks for artifacts that jointly
  expose geometry, topology, constraints, handles, provenance,
  uncertainty, and audit hooks.
\item
  \textbf{Human-in-the-loop operation studies:} evaluate whether
  candidate interpretations, effect diffs, review states, and
  capability-gap explanations improve trust, correction speed, and
  repeated spatial-operation success.
\end{enumerate}

\section{Conclusion}\label{conclusion}

Hylos reframes spatial foundation-model interaction as an operability
problem. Visual 3D generation is not enough. A spatial artifact -
whether an object, assembly, route, scene, or environment - becomes
useful to agents only when it can be inspected, modified, validated,
projected, audited, and committed through a reliable runtime contract.

The reported repair trajectory is scoped but representative because it
stresses the causal half of the broader loop. A visible symptom on a
dependent assembly was traced through scene dependencies to an upstream
placement driver, matched with a supported interaction, and applied as a
validated change. This complements the broader Hylos authoring and
mutation substrate, where desired outcomes are traced to required
topology, supported construction paths, validated scene state, and
consequence-aware updates.

The larger thesis is that the next generation of spatial AI needs a new
artifact contract. Visual 3D generation, scene graphs, CAD kernels,
robotics planners, simulation engines, and agentic interfaces should be
connected by a transaction-safe operability layer. Hylos provides a
staged architecture for getting from today's bounded graph operations to
future model-native spatial artifacts.

\section{References}\label{references}

\begin{enumerate}
\def\labelenumi{\arabic{enumi}.}
\tightlist
\item
  I. Armeni, Z.-Y. He, J. Gwak, A. R. Zamir, M. Fischer, J. Malik, and
  S. Savarese. ``3D Scene Graph: A Structure for Unified Semantics, 3D
  Space, and Camera.'' \emph{IEEE/CVF International Conference on
  Computer Vision (ICCV)}, 2019. \url{https://arxiv.org/abs/1910.02527}
\item
  A. Rosinol, A. Violette, M. Abate, N. Hughes, Y. Chang, J. Shi, A.
  Gupta, and L. Carlone. ``Kimera: From SLAM to Spatial Perception with
  3D Dynamic Scene Graphs.'' \emph{International Journal of Robotics
  Research}, 40(12-14):1510-1546, 2021.
  \url{https://arxiv.org/abs/2101.06894}
\item
  S. Yao, J. Zhao, D. Yu, N. Du, I. Shafran, K. Narasimhan, and Y. Cao.
  ``ReAct: Synergizing Reasoning and Acting in Language Models.''
  \emph{International Conference on Learning Representations (ICLR)},
  2023. \url{https://arxiv.org/abs/2210.03629}
\item
  T. Schick, J. Dwivedi-Yu, R. Dessi, R. Raileanu, M. Lomeli, L.
  Zettlemoyer, N. Cancedda, and T. Scialom. ``Toolformer: Language
  Models Can Teach Themselves to Use Tools.'' \emph{Advances in Neural
  Information Processing Systems (NeurIPS)}, 2023.
  \url{https://arxiv.org/abs/2302.04761}
\item
  M. Ahn et al.~``Do As I Can, Not As I Say: Grounding Language in
  Robotic Affordances.'' \emph{Conference on Robot Learning (CoRL)},
  2022. \url{https://arxiv.org/abs/2204.01691}
\item
  J. Liang et al.~``Code as Policies: Language Model Programs for
  Embodied Control.'' \emph{IEEE International Conference on Robotics
  and Automation (ICRA)}, 2023. \url{https://arxiv.org/abs/2209.07753}
\item
  R. K. Jones, T. Barton, X. Xu, K. Wang, E. Jiang, P. Guerrero, N. J.
  Mitra, and D. Ritchie. ``ShapeAssembly: Learning to Generate Programs
  for 3D Shape Structure Synthesis.'' \emph{ACM Transactions on
  Graphics}, 39(6), 2020. \url{https://arxiv.org/abs/2009.08026}
\item
  M. Deitke et al.~``ProcTHOR: Large-Scale Embodied AI Using Procedural
  Generation.'' \emph{Advances in Neural Information Processing Systems
  (NeurIPS)}, 2022. \url{https://arxiv.org/abs/2206.06994}
\item
  M. Deitke, D. Schwenk, J. Salvador, L. Weihs, O. Michel, E.
  VanderBilt, L. Schmidt, K. Ehsani, A. Kembhavi, and A. Farhadi.
  ``Objaverse: A Universe of Annotated 3D Objects.'' \emph{IEEE/CVF
  Conference on Computer Vision and Pattern Recognition (CVPR)}, 2023.
  \url{https://arxiv.org/abs/2212.08051}
\item
  Y. Yang et al.~``Holodeck: Language Guided Generation of 3D Embodied
  AI Environments.'' \emph{IEEE/CVF Conference on Computer Vision and
  Pattern Recognition (CVPR)}, 2024.
  \url{https://arxiv.org/abs/2312.09067}
\item
  World Labs. ``Marble: A Multimodal World Model.'' Product and
  technical overview, 2025.
  \url{https://www.worldlabs.ai/blog/marble-world-model}
\item
  C. DaSilva. ``Evidence-Grounded Spatial Reasoning with a Prototype
  Semantic-Spatial Research System.'' Internal technical report, 2026.
\end{enumerate}

\end{document}